\documentclass{bmvc2k}

\usepackage{times}
\usepackage{epsfig}
\usepackage{graphicx}
\usepackage{amsmath}
\usepackage{amssymb}
\usepackage{booktabs}
\usepackage{multirow}
\usepackage{mathrsfs}
\usepackage{hyperref}

\title{Learning Embedding of 3D models with Quadric Loss}

\addauthor{Nitin Agarwal}{agarwal@ics.uci.edu}{1}
\addauthor{Sung-eui Yoon}{sungeui@kaist.edu}{2}
\addauthor{M Gopi}{gopi@ics.uci.edu}{1}

\addinstitution{
 Department of Computer Science\\
 University of California, Irvine\\
 California, USA
}
\addinstitution{
 School of Computing\\
 KAIST\\
 South Korea
}

\runninghead{Agarwal ET AL.}{Embedding of 3D models with Quadric Loss}

\newcommand{\Skip}[1] {
}

\begin{document}

\maketitle

\begin{abstract}
Sharp features such as edges and corners play an important role in the perception of 3D models. In order to capture them better, we propose quadric loss, a point-surface loss function, which minimizes the quadric error between the reconstructed points and the input surface. Computation of Quadric loss is easy, efficient since the quadric matrices can be computed apriori, and is fully differentiable, making quadric loss suitable for training point and mesh based architectures. Through extensive experiments we show the merits and demerits of quadric loss. When combined with Chamfer loss, quadric loss achieves better reconstruction results as compared to any one of them or other point-surface loss functions.
\end{abstract}

\section{Introduction}

Following the tremendous success in image classification and detection, 
deep learning based techniques have been widely extended to 3D 
data, opening up numerous 3D applications such as 3D object classification, 
segmentation, shape representation and correspondence finding to name a few. 
In this work we focus on shape representation, particularly on learning a better embedding or shape representation of 3D models using an auto encoder.

Early 3D deep learning techniques 
use 2D and 3D convolution modules to design their network architectures.
Recent techniques extend such convolution modules to handle irregular representations such as
points~\cite{Qi:2017aa,Qi:2017ab} and meshes~\cite{Litany:2017aa, Dai:2018aa,
Tan:2018ac, Ranjan:2018aa}. Together with these architectures, different loss 
functions have been proposed for 3D reconstruction. At a high level, 
they can be classified as being 
between two points (e.g., L1, Earth Mover Distance \cite{Fan:2017aa}) or 
between a point and a surface (e.g., surface loss \cite{Yu:2018ab}). Among these loss functions, 
Chamfer loss \cite{Fan:2017aa,Yu:2018ab} has been widely used for reconstructing 3D models.

\begin{figure*}[!t] 
\centering
\begin{center}
\includegraphics[width=0.99\textwidth,height=3cm]{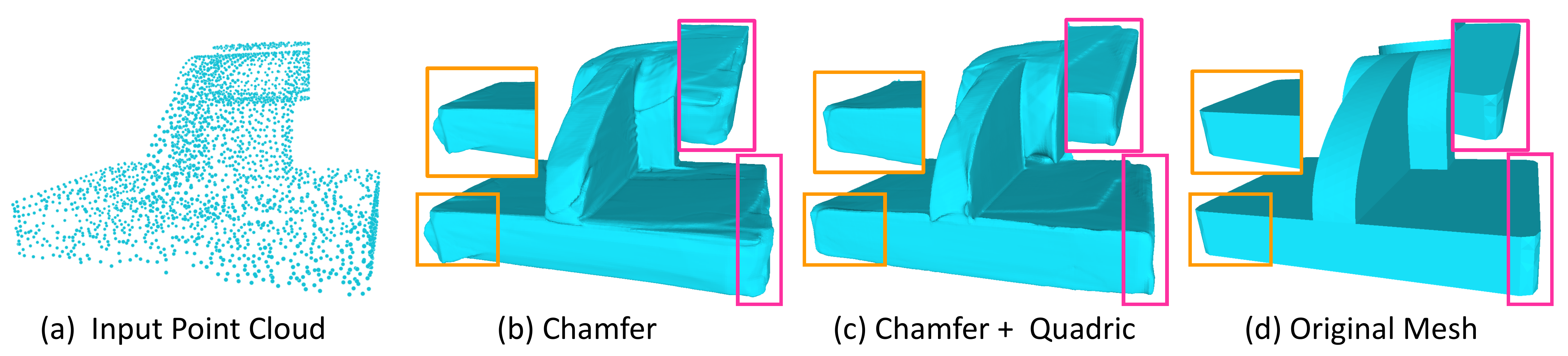}
\end{center}
\caption{(a) Input point cloud reconstructed using an auto-encoder network with (b) Chamfer loss alone and (c) Chamfer + Quadric loss. Reconstructed meshes are generated using Poisson surface reconstruction on output point cloud.}
\label{fig:teaserImg}
\end{figure*}

While these loss functions work well in maintaining the overall 
structure of the 3D model, they do not preserve high-frequency information such 
as edges and corners. To address this issue, we propose a novel loss function, quadric loss, 
for preserving such detailed structures. Inspired by mesh simplification techniques, 
quadric loss is defined as the sum of squared distances between a reconstructed point and
planes defined by triangles incident to its corresponding point in the input mesh.  
Intuitively, the quadric loss penalizes the displacement of points along the normal direction of those
planes, maintaining sharp edges and corners (Fig. \ref{fig:teaserImg}). 

To demonstrate the benefits of quadric loss, we conduct experiments
with 3D CAD models, and compare various loss functions both qualitatively and quantitatively. 
Overall, we find that the combination of Chamfer and our quadric loss shows the
best result, since Chamfer loss maintains the overall structure and point distribution, while the quadric loss preserves sharp features. 

To summarize, our main contributions in this work are:
\begin{itemize}
    \item We propose a new point-surface loss named \emph{quadric loss}, which preserves sharp features such as corners and edges in the reconstructed models. It is fast, easy to compute and is fully differentiable. It does not introduce any hyperparameters and can be used with most existing point/mesh based architectures without modification.
    \item We evaluate our loss function extensively and also provide its geometric interpretation.
    \item We compare our quadric loss with other point-surface loss functions and the popular Chamfer loss and discuss in detail the merit and demerit of each.  
\end{itemize}

\section{Related Works}

\subsection{Learning Shape Representation} 

There is a rich literature for learning compact 3D shape representations using
deep learning techniques. Prior works \cite{Su:2015aa, Li:2015aa,
Girdhar:2016aa, Wu:2016aa} have used image and voxel based representations of 3D
models to learn a discriminative representation for the task of 3D object
recognition, classification and generation. Although their structured
representations facilitate the use of traditional 2D and 3D convolution, they
are not readily available for
handling complex and high resolution models. On the other hand, part-based
approaches \cite{Li:2017ab, Nash:2017aa, Wu:2018aa} can produce shapes with
complex structures, but the level of detail is restricted to the components and
primitives used. 

\begin{figure*}[t] 
\centering
\begin{center}
\includegraphics[width=0.99\textwidth,height=6cm]{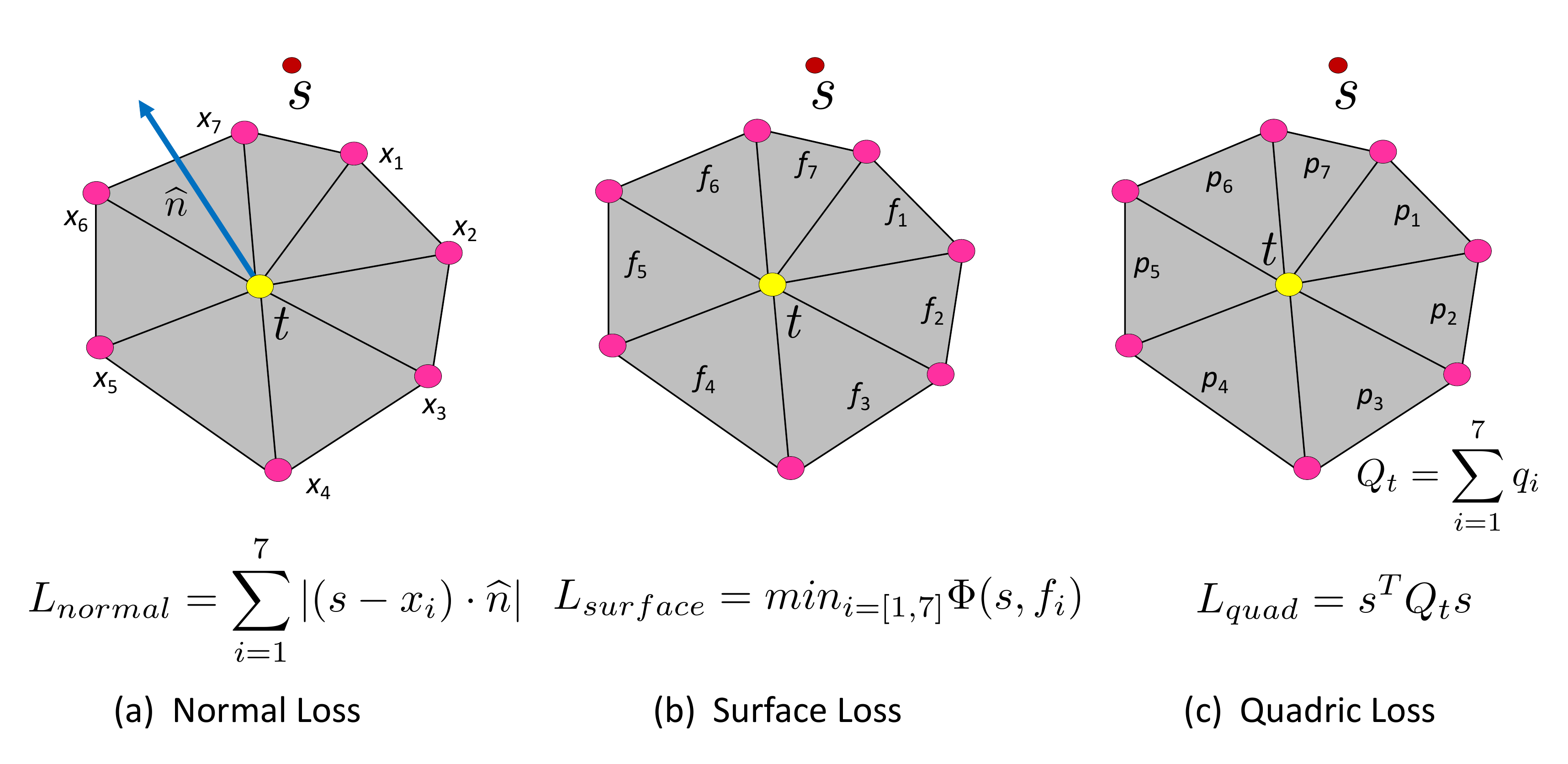}
\end{center}
	\caption{Computation of point-surface losses: Let the reconstructed point $s$
	correspond to the point $t$ in the input mesh. (a) Normal loss computes
	the inner product between the edge formed by $s$ and $x_i$
	and the ground truth normal vector $\widehat{n}$ at $t$; (b) Surface
	loss computes the point-triangle distance $\Phi$ between $s$ and $f$,
	where $f$ represents a triangle and \emph{not a plane}, and takes the minimum of them with different triangles; (c) Quadric
	loss (our contribution) computes the sum of the square of the distance
	between $s$ and each of the plane $p$ ($p=[a,b,c,d]^T$) formed by the triangle incident at
	$t$ using the quadric matrix $q_i$ which is computed as $q_i = p_ip_i^T$. Please see Eq. \ref{equ:quadError} for more details.}
	\label{fig:surfaceLoss}
\end{figure*}

Recently, convolution has been extended to more unstructured representations
like 3D point datasets and meshes. PointNet \cite{Qi:2017aa} and PointNet++
\cite{Qi:2017ab} have been widely used as an encoder to achieve superior
performance on various tasks such as object classification \cite{Qi:2017aa,
Wang:2018ab}, segmentation \cite{Qi:2017aa, Wang:2018ab}, point set generation
\cite{Achlioptas:2017aa,Yang:2018aa}, shape correspondence
\cite{Groueix:2018aa} etc. Mesh based networks have also been used to learn
embeddings for shape completion \cite{Litany:2017aa, Dai:2018aa} and shape
deformation \cite{Tan:2018ac, Ranjan:2018aa}.

Since point and mesh based representations, when compared to voxel-based representations, are light-weight, flexible in terms of reconstructing complex models and scale well to high resolution models, we propose a loss function which can be used by such networks to further enhance the embedding and reconstruction quality of 3D models.

\noindent
\subsection{3D Reconstruction Losses} 
Losses commonly used with point and mesh based networks for 3D reconstruction can be broadly classified into two categories - between two points or between a point and a surface.

\noindent
\textbf{Point based Loss:} Point based loss functions compute the dissimilarity
between two pointset distributions. Losses like L1 \cite{Dai:2018aa} and L2
\cite{Groueix:2018aa,Litany:2017aa} require both one-to-one correspondence and
the cardinality of the two pointsets to be the same. Earth movers distance
(EMD) or Wasserstein metric \cite{Fan:2017aa} is similar to these losses as it
requires the input cardinality between pointsets to be the same. It solves an
optimization problem where it computes a bijection between the two pointsets.
However, a major drawback of EMD is that it is both memory and compute
intensive, hence is usually approximated \cite{Fan:2017aa}. Chamfer distance
(CD) \cite{Fan:2017aa,Wang:2018aa,Groueix:2018ab}, which has become a standard
for reconstructing 3D objects, computes the shortest distance of each point 
in one pointset to the other pointset. This distance is computed in both directions. 
It does not require the cardinality of the input points to be the same nor does it require
any one-to-one correspondence. Although CD works well at recovering the overall
structure, it does not preserve sharp features like corners and edges, and often results in
collapse of smaller structures \cite{Dai:2018aa}.

\noindent
\textbf{Point-Surface based Loss:} Point-surface based loss functions try to
minimize the distance between the output reconstructed point and the input
surface. Yu et al. \cite{Yu:2018ab} propose \emph{surface-loss}
(point-triangle), which computes the minimum of the shortest distances between
an output point and each triangle in a subset of triangles defining the input mesh
(Fig \ref{fig:surfaceLoss}b). Similar to \emph{surface-loss}, Yu et al. \cite{Yu:2018ab} also propose
\emph{edge-loss}, which requires the edges in the input model to be manually annotated.
Wang et al. \cite{Wang:2018aa} propose \emph{normal-loss} to incorporate 
higher-order features in their reconstruction. It minimizes the inner product 
of the edge formed from the output point and the neighbours of the corresponding 
input point with its normal vector. In other words, it requires the edges between 
the output point and the neighbours of the corresponding input point to be orthogonal 
to the ground truth normal vector (Fig \ref{fig:surfaceLoss}a).

Departing from these prior approaches, we propose a new point-surface based
loss function named \emph{quadric loss}, which encourages sharp corners and edges
to be preserved in the output reconstruction (Fig \ref{fig:teaserImg} and Fig. \ref{fig:single_loss}). Unlike edge-loss, quadric loss does not require the edges to be annotated in the input models.  
Quadric loss minimizes the distances between the output point and the planes defined by 
the triangles incident to its corresponding point in the input mesh (Fig \ref{fig:surfaceLoss}c). 
It is fast and easy to compute as oppose to surface loss for which one needs to consider all the seven cases, as the point which minimizes the point-triangle distance can be on the 3 edges, 3 vertices or inside the triangle \cite{Yu:2018ab}. Quadric loss is also differentiable making it amiable for training via back propagation.

\section{Quadric Loss}

\begin{figure*}[!b] 
\centering
\begin{center}
\includegraphics[width=0.7\textwidth,height=5cm]{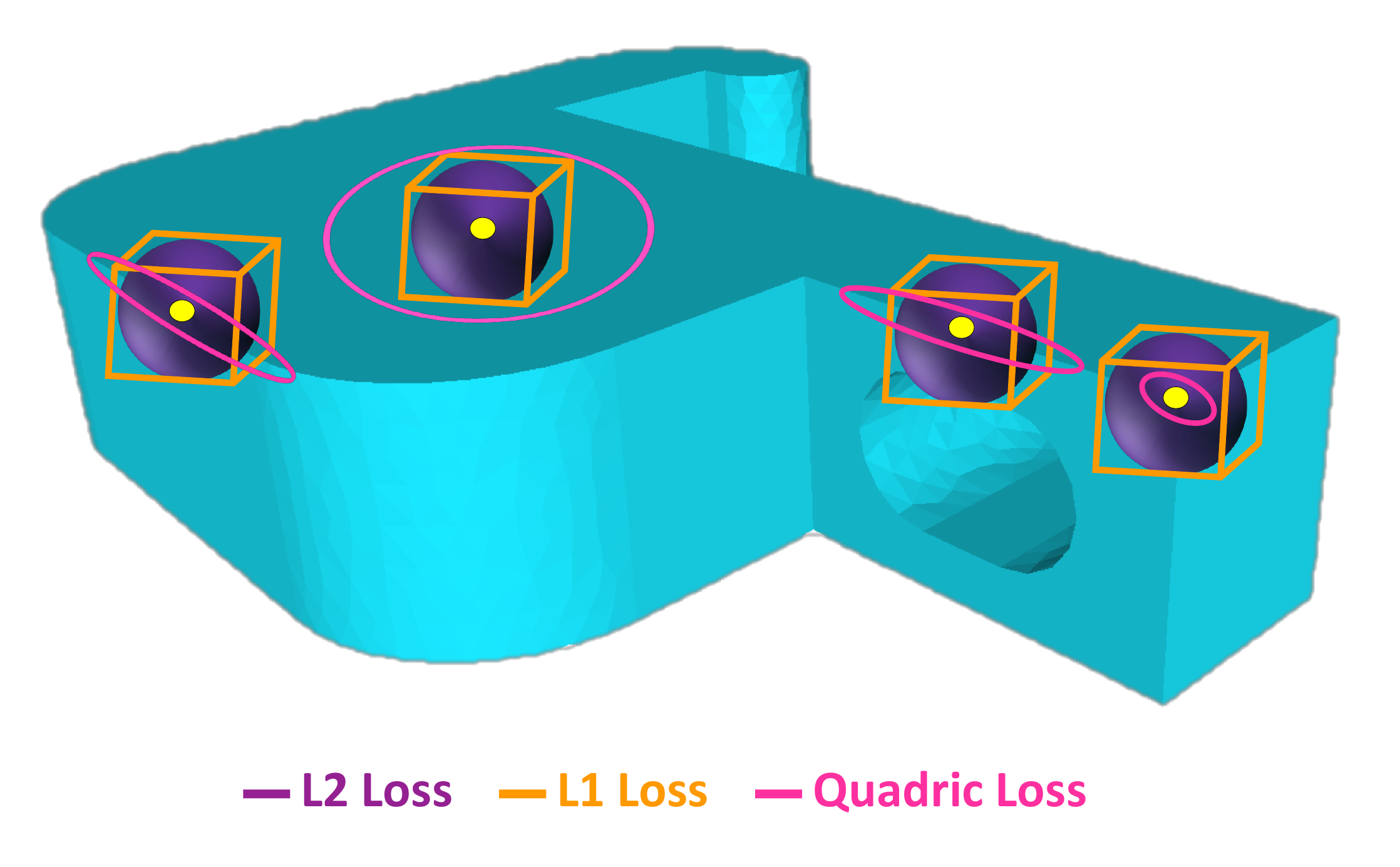}
\end{center}
\caption{Geometric Interpretation of quadric loss: Quadric loss is an
\emph{ellipsoidal loss} and it penalizes the reconstructed points more in the
normal direction. Here we show the iso-error envelope of Quadric, L1 and L2.
For illustration purposes, we draw iso-errors in 2D on few points (yellow) on the
input surface. Points lying on flat planes would ideally have ellipsoids with 0 minor axis 
and $\infty$ major axes lengths, i.e the reconstructed points can be placed anywhere on the plane.
Note that the ellipsoid for points on sharp features like corners is very small compared to
L1 and L2, ensuring the reconstructed points to preserve such features.}
\label{fig:quadricLoss}
\end{figure*}

Quadric error metric was originally proposed for mesh simplification \cite{Ronfard:1996aa,
Garland:1997aa}, i.e, the task of reducing of a mesh with a high polygon count
to a low polygon count while preserving as much visual geometric detail as possible.
Quadric error computes the squared distance between a point and a plane in ${\rm I\!R}^3$.
Inspired by this quadric error metric, we propose \emph{quadric loss}, a point-surface loss function,
which penalizes the reconstructed points in the normal direction, thereby preserving sharp
edges and corners in the output reconstruction as shown in Figure \ref{fig:teaserImg}.

\noindent
\textbf{Background:} 
Let a point $s$ be represented in homogeneous coordinates $[x,y,z,1]^T$, and a plane $p$ be represented as $[a,b,c,d]^T$ where $a^2+b^2+c^2=1$. The distance of $s$ from $p$ is given by $ax+by+cz+d$, which can be computed as $p^Ts$. The square of the distance of $s$ from $p$ is given by

\begin{equation}
(p^Ts)^2 = (p^Ts)(p^Ts) = s^T(pp^T)s=s^TQ_ps,
\label{equ:quadError}
\end{equation}

\noindent
where $Q_p$ is a symmetric matrix called the quadric matrix \cite{Garland:1997aa}, determined only by the plane and not by the point. Given a set of planes $p_1,p_2,...p_k$, the sum of the squared distance of $s$ from this set of planes is given by

\begin{equation}
\Delta(s) = \sum{s^TQ_is} = s^T (\sum{Q_i}) s = s^TQs,  \hspace{5mm} where \hspace{5mm} Q = \sum{Q_i}.  
\label{equ:quadError2}
\end{equation}

\noindent
It should be clear that in a mesh, the quadric error of a vertex $s$ from the planes defined by the triangles incident on $s$ is zero. 

\noindent
\textbf{Computation:} Given an input mesh $\mathcal{M}$ with $\mathcal{V}_{in} \in {\rm I\!R}^{Nx3}$ vertices and a set of reconstructed points $\mathcal{V}_{out} \in {\rm I\!R}^{Nx3}$, let $s$ be a reconstructed point corresponding to input vertex $t$. We want $s$ to be on all the triangles incident on $t$ just as $t$ is on those planes in the input mesh. So the quadric error of $s$ from the planes defined by the triangles incident on its corresponding point $t$, namely $s^TQs$, has to be minimized. We call $s^TQs$ as the quadric loss, which we compute between $\mathcal{V}_{in}$ and $\mathcal{V}_{out}$ as the following:

\begin{equation}
L_{quad} = \frac{1}{N}\sum_{\substack{s \in \mathcal{V}_{out} \\ t \in \mathcal{V}_{in}}}s^TQ_ts.
\label{equ:quadricLoss}
\end{equation}

\noindent
\textbf{Geometric Interpretation:} The iso-value surfaces $s^TQs$ defined by the quadric matrix Q at the input vertex $t$, represents a family of ellipsoids centered at $t$, for which one of the three axes corresponds to the normal vector of the surface at $t$. The length of the other two axes are inversely proportional to the curvature of the surface in those directions. For example, in a planar region, the length of the ellipsoidal axes is infinity along the plane and zero along the normal vector direction. In other words, the reconstructed point can be anywhere on the plane, but any displacement along the normal vector direction will introduce more quadric loss. For vertices along a sharp, straight edge of a 3D model, the quadric error ellipsoid will have infinite length along the edge and zero length for the other two axes. In other words, the reconstructed point can be placed anywhere along the straight edge for the quadric error to be still zero, but any displacement away from the edge will incur a loss. A similar argument holds for a pointed corner of a 3D model. The quadric ellipsoid will be very small, restricting the freedom of placement of the reconstructed point as shown in Figure \ref{fig:quadricLoss}. Hence, unlike Chamfer and L2 loss which are \emph{spherical losses} - points equidistant from the input vertex have equal loss, quadric loss is an \emph{ellipsoidal loss} which penalizes displacement of points more in the normal direction.

\section{Experiments}

In this section, we present the results of training an auto-encoder with various
point-surface loss functions. Specifically, we compare our quadric loss with
surface loss \cite{Yu:2018ab} and normal loss \cite{Wang:2018aa}
for the task of shape reconstruction. We analyze the reconstruction results 
both qualitatively and quantitatively, and also compare our proposed loss with 
the popular Chamfer loss.

\begin{figure*}[!t] 
\centering
\begin{center}
\includegraphics[width=0.99\textwidth,height=7cm]{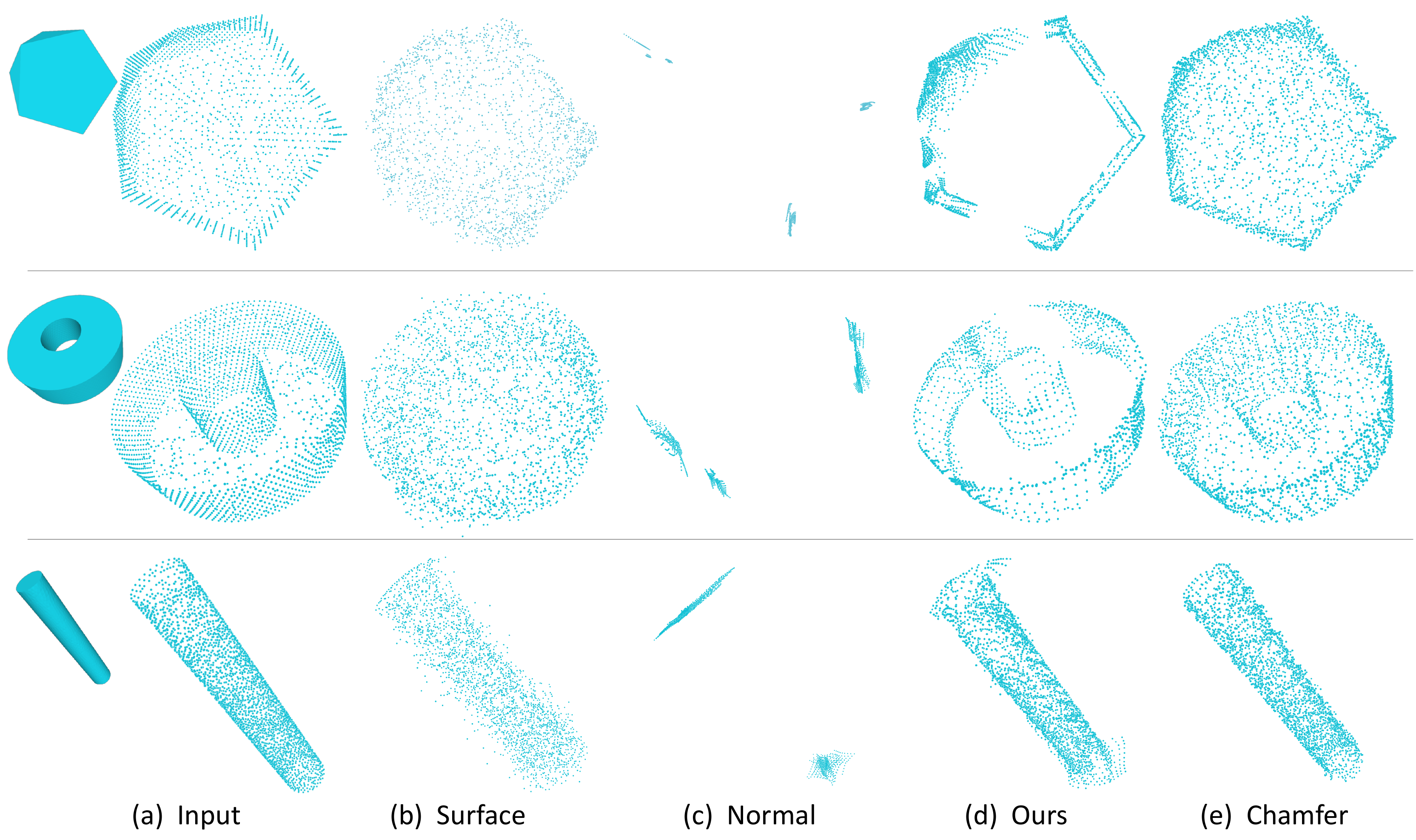}
\end{center}
\caption{Effect of Point-Surface loss: Reconstruction results (2500 points) on
example 3D models from the test set with different loss functions. In comparison to 
Chamfer focusing on preserving the input point distribution,
our quadric loss encourages points to be on edges and corners. On flat planes
(like the top and bottom faces of the cylinders in the bottom row) reconstructed points minimize the quadric error by lying on the plane, but can be outside the ground truth model. Such artifacts can be avoided by
the combination of quadric and Chamfer loss (top row of Fig \ref{fig:combine_loss}).}
\label{fig:single_loss}
\end{figure*}

\subsection{Dataset} 

To train the autoencoder, we use the recently published ABC dataset
\cite{Koch:2018aa}. Although this dataset contains more than 1 million high
quality CAD models of mechanical parts, we randomly selected 5000 CAD models
for our experiment. The reason of using ABC dataset over other 3D shape
repositories like ShapeNet \cite{shapenet2015} and ModelNet40 \cite{Wu:2015aa}
is the presence of sharp edges and corners, which are commonly found in
mechanical parts (Fig. \ref{fig:combine_loss}). As some of the models comprised of
multiple disconnected components, we separated each model into its connected 
components such that each model has a single mesh. This increased our dataset 
size to 8064 models. We also simplified the models using Q-slim \cite{Garland:1997aa} 
to reduce the vertex count to 2500 vertices, and centered and normalized them
to a unit sphere. We randomly split the data to get a distribution of 90\% for
training and 10\% for testing.

\subsection{Network \& Implementation Details}

Although Quadric Loss can potentially be used with any point or mesh based
network, we use an auto-encoder based network and analyze the reconstruction 
quality during shape reconstruction. We use the encoder from Dynamic Graph CNN (DGCNN) \cite{Wang:2018ab},
which performs convolution over $k$-nearest neighbours in the feature space at
every layer and is currently the state of the art for point cloud analysis.
Specifically, we use the classification architecture without the spatial
transformer and the fully connected layers to encode a point cloud of 2500
vertices into a latent vector dimension of 1024.

For the decoder we use AtlasNet \cite{Groueix:2018ab}, which takes in the 1024 embedding
from the DGCNN encoder and generates an output surface using $N$ learnt parameterizations. 
We follow the same training strategy as AtlasNet, which is to sample the learned
parameterizations at every training step to avoid over-fitting. For all the
experiments in this paper, we use this auto-encoder architecture with $k$ = 20,
$N$ = 25 and an output point cloud size of 2500.

In order to compare the three point-surface loss functions, we train 4 networks - 
one with CD + surface loss, one with CD + normal loss, one with CD + quadric loss 
and one with CD alone. To compute the three losses (surface, normal and quadric), we use 
the correspondences found from Chamfer distance. For all the experiments we use 
Adam \cite{Kingma:2014aa} optimizer with a batch size of 16. The learning rate 
was set to 0.001 for all losses except the networks trained with quadric loss for which we 
found a slower learning rate of 0.0001 to be most effective. All learning rates 
were multiplied by 0.8 every 100 epochs. For a fair comparison we train all the 
networks to the same number of epochs and we also ensure that the total loss 
in each network is an equal contribution of both the loss functions by weighting 
the terms appropriately. All the code was implemented in Pytorch and training 
was performed on NVIDIA TITAN Xp GPU. 

\begin{table}[!t]
  \centering
  \begin{tabular}{lcccc}
    \toprule
    \multirow{2}{*}{Losses}& \multicolumn{2}{c}{CD} & \multicolumn{2}{c}{Metro} \\
    \cmidrule(lr){2-3} \cmidrule(lr){4-5}
    &median &max &median &max\\
    \midrule
    Normal loss &397.09 &1750.6 &10.65 &28.38 \\
    Surface loss &21.86 &398.85 &6.11 &24.93 \\
    Quadric loss  &9.44 &217.5 &3.18 &20.80\\
    Chamfer loss &\textbf{1.97} &40.87 &3.13 &19.08 \\
    \midrule
    Normal + Chamfer loss &2.97 &39.83 &3.38 &19.21 \\
    Surface + Chamfer loss &2.23 &37.04 &3.16 &18.87 \\
    Quadric + Chamfer loss &2.21 &\textbf{36.78} &\textbf{2.96} &\textbf{18.80} \\
    \bottomrule
  \end{tabular}
  \vspace{1mm}
\caption{3D reconstruction results on models from the test set. We compare different loss functions using Chamfer distance (CD), computed on 2500 points, multiplied by $10^3$ and Metro error \cite{Cignoni:1998aa}, multiplied by 10. Among all four losses, Chamfer loss best preserves the overall structure and point distribution which is reflected in its low CD and Metro values. Quadric loss preserves sharp edges and corners (Fig. \ref{fig:single_loss}) but has a higher CD when compared to Chamfer loss. Combining quadric with Chamfer achieves best results.}
\label{tab:numerical}
\end{table}

\subsection{Evaluation Metric}

To evaluate the quality of the reconstructed shapes, we compare it with the
ground truth shapes using two criteria. First, we compare the Chamfer distance
(CD) \cite{Fan:2017aa} between the input and output point clouds. CD alone is
a necessary but not a sufficient condition for a good reconstruction; CD can be minimized by
assigning just one point in one point cloud to a cluster of points in the other
point cloud. Hence, we also compare the Metro error between the input and output
meshes using the publicly available software \cite{Cignoni:1998aa}. Simply put,
it computes the Euclidean distance between two meshes by sampling points on
them. We report the maximum distance between the two meshes because outliers dictate the visual quality 
and fidelity of the reconstructed mesh.

\begin{figure*}[!t] 
\centering
\begin{center}
\includegraphics[width=0.99\textwidth,height=9cm]{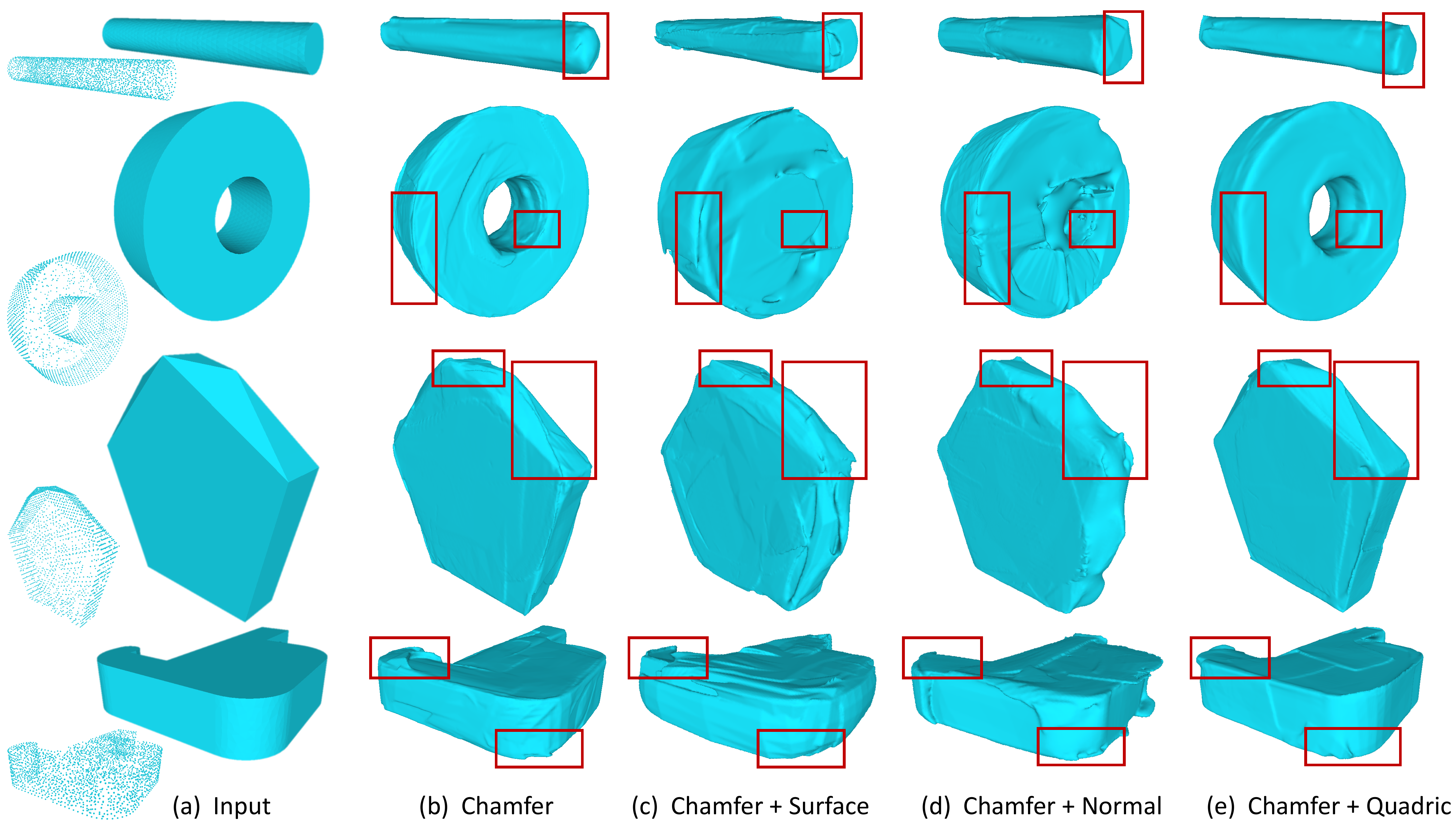}
\end{center}
\caption{Reconstruction results of 3D models from the test set. To obtain a mesh from the reconstructed point clouds, we follow a similar procedure as AtlasNet \cite{Groueix:2018ab}. i.e. we shoot rays at the model from infinity to obtain a dense sample of points followed by Poisson surface reconstruction (PSR) \cite{Kazhdan:2006aa}. Chamfer loss when added to surface, normal and quadric loss improves the reconstruction result as compared to them individually. Note, sharp edges and corners are achieved with quadric and Chamfer togather. For more qualitative results please see the supplementary material.}
\label{fig:combine_loss}
\end{figure*}

\subsection{Shape Reconstruction}

We evaluate the learnt embedding by analyzing the reconstruction quality of the 3D 
models. We report the quantitative results in Table \ref{tab:numerical} 
where the results are from computing the median and the maximum values of all models
in the test set. 

\noindent
\textbf{Reconstruction without Chamfer:} In order to study the effect of various point-surface losses, we train the auto encoder with surface, normal, and quadric losses independently without chamfer loss. We report the qualitative results in Figure \ref{fig:single_loss}. As compared to surface and normal loss, quadric loss alone reconstructs the models much better. Through our extensive experiments we find training of normal loss also to be much difficult as compared to surface and quadric loss. i.e. it does not converge. Quadric loss in comparison to surface loss preserves the sharp features better while surface loss is able to achieve better point distribution. As surface loss computes the closest triangle to the reconstructed point and minimizes that distance, it is difficult for it to reconstruct sharp features like edges and corners. 

As quadric loss encourages more reconstructed points to lie along the edges and corners, it has a higher CD than network trained with Chamfer loss alone. Also, notice the small difference in Metro error between Chamfer and quadric loss. This is because Metro error is computed by sampling the  meshes and not the points (like CD). Also, quadric loss with AtlasNet decoder is able to reconstruct the patches ($N$ learnt parameterizations) close to the input surface. This demonstrates that Metro error does not care about the point distribution as long as the output mesh surface is close to the input mesh surface. Hence, a 
good reconstruction should preserve both the point distribution (low CD) and overall structure (low Metro). 

\noindent
\textbf{Reconstruction with Chamfer:} Chamfer loss when added improves reconstructions based on surface, normal, or quadric losses (Fig. \ref{fig:combine_loss}). Quadric with chamfer achieves the best reconstruction results overall. Addition of quadric to Chamfer loss further reduces the maximum CD from 40.87 to 36.78. This is because as Chamfer loss tries to preserve the point distribution, quadric loss tries to preserve sharp features like edges and corners. Hence, models reconstructed using both quadric and Chamfer enjoy best of both worlds - sharp features and good point distribution.

\section{Conclusion}

In this work we propose a new point-surface loss function, named quadric loss, which penalizes the displacement of points in the normal direction thereby preserving sharp features like edges and corners in the reconstructed models. Quadric loss is easy to compute, fully differentiable and can be integrated into most point and mesh based architectures. Quadric loss can also successfully reconstruct models having no sharp features. However, as quadric loss is an \emph{ellipsoidal loss}, it cannot preserve the input point distribution. For points on the planar faces of a surface, since the quadric loss is zero anywhere on the plane, the reconstructed points may lie outside the extents of the planar face. Hence, quadric loss should always be accompanied with a \emph{spherical loss} like chamfer loss which preserves the input point distribution. Note that Chamfer has its own weakness; its value could be minimized by assigning one point to a cluster of points.  Depending on the application, these two losses could be weighted appropriately. Since Chamfer and quadric loss functions complement each other, combining these two loss functions achieve better embedding than using any one of them. 

\noindent
\textbf{Acknowledgements.} Prof. Yoon was supported in part by NRF-2017M3C4A7066317 and NRF/MSIT 
(No. 2019R1A2C3002833).

\bibliography{refs}

\begin{thebibliography}{29}
\providecommand{\natexlab}[1]{#1}
\providecommand{\url}[1]{\texttt{#1}}
\expandafter\ifx\csname urlstyle\endcsname\relax
  \providecommand{\doi}[1]{doi: #1}\else
  \providecommand{\doi}{doi: \begingroup \urlstyle{rm}\Url}\fi

\bibitem[Achlioptas et~al.(2018)Achlioptas, Diamanti, Mitliagkas, and
  Guibas]{Achlioptas:2017aa}
Panos Achlioptas, Olga Diamanti, Ioannis Mitliagkas, and Leonidas Guibas.
\newblock Learning representations and generative models for 3d point clouds.
\newblock In \emph{ICML}, 2018.

\bibitem[Chang et~al.(2015)Chang, Funkhouser, Guibas, Hanrahan, Huang, Li,
  Savarese, Savva, Song, Su, Xiao, Yi, and Yu]{shapenet2015}
Angel~X. Chang, Thomas Funkhouser, Leonidas Guibas, Pat Hanrahan, Qixing Huang,
  Zimo Li, Silvio Savarese, Manolis Savva, Shuran Song, Hao Su, Jianxiong Xiao,
  Li~Yi, and Fisher Yu.
\newblock {ShapeNet: An Information-Rich 3D Model Repository}.
\newblock Technical Report arXiv:1512.03012 [cs.GR], Stanford University ---
  Princeton University --- Toyota Technological Institute at Chicago, 2015.

\bibitem[Cignoni et~al.(1998)Cignoni, Rocchini, and Scopigno]{Cignoni:1998aa}
Paolo Cignoni, Claudio Rocchini, and Roberto Scopigno.
\newblock Metro: Measuring error on simplified surfaces.
\newblock In \emph{Computer Graphics Forum}, 1998.

\bibitem[Dai and Nie{\ss}ner(2018)]{Dai:2018aa}
Angela Dai and Matthias Nie{\ss}ner.
\newblock Scan2mesh: From unstructured range scans to 3d meshes.
\newblock \emph{arXiv preprint arXiv:1811.10464}, 2018.

\bibitem[Fan et~al.(2017)Fan, Su, and Guibas]{Fan:2017aa}
Haoqiang Fan, Hao Su, and Leonidas Guibas.
\newblock A point set generation network for 3d object reconstruction from a
  single image.
\newblock In \emph{CVPR}, 2017.

\bibitem[Garland and Heckbert(1997)]{Garland:1997aa}
Michael Garland and Paul~S Heckbert.
\newblock Surface simplification using quadric error metrics.
\newblock In \emph{SIGGRAPH}, 1997.

\bibitem[Girdhar et~al.(2016)Girdhar, Fouhey, Rodriguez, and
  Gupta]{Girdhar:2016aa}
Rohit Girdhar, David~F Fouhey, Mikel Rodriguez, and Abhinav Gupta.
\newblock Learning a predictable and generative vector representation for
  objects.
\newblock In \emph{ECCV}, 2016.

\bibitem[Groueix et~al.(2018{\natexlab{a}})Groueix, Fisher, Kim, Russell, and
  Aubry]{Groueix:2018aa}
Thibault Groueix, Matthew Fisher, Vladimir~G Kim, Bryan~C Russell, and Mathieu
  Aubry.
\newblock Shape correspondences from learnt template-based parametrization.
\newblock In \emph{ECCV}, 2018{\natexlab{a}}.

\bibitem[Groueix et~al.(2018{\natexlab{b}})Groueix, Fisher, Kim, Russell, and
  Aubry]{Groueix:2018ab}
Thibault Groueix, Matthew Fisher, Vladimir~G Kim, Bryan~C Russell, and Mathieu
  Aubry.
\newblock Atlasnet: A papier approach to learning 3d surface generation.
\newblock In \emph{CVPR}, 2018{\natexlab{b}}.

\bibitem[Kazhdan et~al.(2006)Kazhdan, Bolitho, and Hoppe]{Kazhdan:2006aa}
Michael Kazhdan, Matthew Bolitho, and Hugues Hoppe.
\newblock Poisson surface reconstruction.
\newblock In \emph{SGP}, 2006.

\bibitem[Kingma and Ba(2014)]{Kingma:2014aa}
Diederik~P Kingma and Jimmy Ba.
\newblock Adam: A method for stochastic optimization.
\newblock \emph{arXiv preprint arXiv:1412.6980}, 2014.

\bibitem[Koch et~al.(2019)Koch, Matveev, Jiang, Williams, Artemov, Burnaev,
  Alexa, Zorin, and Panozzo]{Koch:2018aa}
Sebastian Koch, Albert Matveev, Zhongshi Jiang, Francis Williams, Alexey
  Artemov, Evgeny Burnaev, Marc Alexa, Denis Zorin, and Daniele Panozzo.
\newblock Abc: A big cad model dataset for geometric deep learning.
\newblock In \emph{CVPR}, 2019.

\bibitem[Li et~al.(2017)Li, Xu, Chaudhuri, Yumer, Zhang, and Guibas]{Li:2017ab}
Jun Li, Kai Xu, Siddhartha Chaudhuri, Ersin Yumer, Hao Zhang, and Leonidas
  Guibas.
\newblock Grass: Generative recursive autoencoders for shape structures.
\newblock \emph{ACM Transactions on Graphics (TOG)}, 2017.

\bibitem[Li et~al.(2015)Li, Su, Qi, Fish, Cohen-Or, and Guibas]{Li:2015aa}
Yangyan Li, Hao Su, Charles~Ruizhongtai Qi, Noa Fish, Daniel Cohen-Or, and
  Leonidas~J Guibas.
\newblock Joint embeddings of shapes and images via cnn image purification.
\newblock \emph{ACM Transactions on Graphics (TOG)}, 2015.

\bibitem[Litany et~al.(2018)Litany, Bronstein, Bronstein, and
  Makadia]{Litany:2017aa}
Or~Litany, Alex Bronstein, Michael Bronstein, and Ameesh Makadia.
\newblock Deformable shape completion with graph convolutional autoencoders.
\newblock In \emph{CVPR}, 2018.

\bibitem[Nash and Williams(2017)]{Nash:2017aa}
Charlie Nash and Chris~KI Williams.
\newblock The shape variational autoencoder: A deep generative model of
  part-segmented 3d objects.
\newblock \emph{Computer Graphics Forum}, 2017.

\bibitem[Qi et~al.(2017{\natexlab{a}})Qi, Su, Mo, and Guibas]{Qi:2017aa}
Charles~R Qi, Hao Su, Kaichun Mo, and Leonidas~J Guibas.
\newblock Pointnet: Deep learning on point sets for 3d classification and
  segmentation.
\newblock In \emph{CVPR}, 2017{\natexlab{a}}.

\bibitem[Qi et~al.(2017{\natexlab{b}})Qi, Yi, Su, and Guibas]{Qi:2017ab}
Charles~Ruizhongtai Qi, Li~Yi, Hao Su, and Leonidas~J Guibas.
\newblock Pointnet++: Deep hierarchical feature learning on point sets in a
  metric space.
\newblock In \emph{NeurIPS}, 2017{\natexlab{b}}.

\bibitem[Ranjan et~al.(2018)Ranjan, Bolkart, Sanyal, and Black]{Ranjan:2018aa}
Anurag Ranjan, Timo Bolkart, Soubhik Sanyal, and Michael~J Black.
\newblock Generating 3d faces using convolutional mesh autoencoders.
\newblock In \emph{ECCV}, 2018.

\bibitem[Ronfard and Rossignac(1996)]{Ronfard:1996aa}
R{\'e}mi Ronfard and Jarek Rossignac.
\newblock Full-range approximation of triangulated polyhedra.
\newblock \emph{Computer Graphics Forum}, 1996.

\bibitem[Su et~al.(2015)Su, Maji, Kalogerakis, and Learned-Miller]{Su:2015aa}
Hang Su, Subhransu Maji, Evangelos Kalogerakis, and Erik Learned-Miller.
\newblock Multi-view convolutional neural networks for 3d shape recognition.
\newblock In \emph{ICCV}, 2015.

\bibitem[Tan et~al.(2018)Tan, Gao, Lai, Yang, and Xia]{Tan:2018ac}
Qingyang Tan, Lin Gao, Yu-Kun Lai, Jie Yang, and Shihong Xia.
\newblock Mesh-based autoencoders for localized deformation component analysis.
\newblock In \emph{AAAI}, 2018.

\bibitem[Wang et~al.(2018{\natexlab{a}})Wang, Zhang, Li, Fu, Liu, and
  Jiang]{Wang:2018aa}
Nanyang Wang, Yinda Zhang, Zhuwen Li, Yanwei Fu, Wei Liu, and Yu-Gang Jiang.
\newblock Pixel2mesh: Generating 3d mesh models from single rgb images.
\newblock In \emph{ECCV}, 2018{\natexlab{a}}.

\bibitem[Wang et~al.(2018{\natexlab{b}})Wang, Sun, Liu, Sarma, Bronstein, and
  Solomon]{Wang:2018ab}
Yue Wang, Yongbin Sun, Ziwei Liu, Sanjay~E Sarma, Michael~M Bronstein, and
  Justin~M Solomon.
\newblock Dynamic graph cnn for learning on point clouds.
\newblock \emph{arXiv preprint arXiv:1801.07829}, 2018{\natexlab{b}}.

\bibitem[Wu et~al.(2016)Wu, Zhang, Xue, Freeman, and Tenenbaum]{Wu:2016aa}
Jiajun Wu, Chengkai Zhang, Tianfan Xue, Bill Freeman, and Josh Tenenbaum.
\newblock Learning a probabilistic latent space of object shapes via 3d
  generative-adversarial modeling.
\newblock In \emph{NeurIPS}, 2016.

\bibitem[Wu et~al.(2018)Wu, Wang, Lin, Lischinski, Cohen-Or, and
  Huang]{Wu:2018aa}
Zhijie Wu, Xiang Wang, Di~Lin, Dani Lischinski, Daniel Cohen-Or, and Hui Huang.
\newblock Structure-aware generative network for 3d-shape modeling.
\newblock \emph{arXiv preprint arXiv:1808.03981}, 2018.

\bibitem[Wu et~al.(2015)Wu, Song, Khosla, Yu, Zhang, Tang, and Xiao]{Wu:2015aa}
Zhirong Wu, Shuran Song, Aditya Khosla, Fisher Yu, Linguang Zhang, Xiaoou Tang,
  and Jianxiong Xiao.
\newblock 3d shapenets: A deep representation for volumetric shapes.
\newblock In \emph{CVPR}, 2015.

\bibitem[Yang et~al.(2018)Yang, Feng, Shen, and Tian]{Yang:2018aa}
Yaoqing Yang, Chen Feng, Yiru Shen, and Dong Tian.
\newblock Foldingnet: Point cloud auto-encoder via deep grid deformation.
\newblock In \emph{CVPR}, 2018.

\bibitem[Yu et~al.(2018)Yu, Li, Fu, Cohen-Or, and Heng]{Yu:2018ab}
Lequan Yu, Xianzhi Li, Chi-Wing Fu, Daniel Cohen-Or, and Pheng-Ann Heng.
\newblock Ec-net: an edge-aware point set consolidation network.
\newblock In \emph{ECCV}, 2018.

\end{thebibliography}
\end{document}